\documentclass[conference]{IEEEtran}

\usepackage{amsmath,amssymb}
\usepackage{graphicx}
\usepackage{url}
\usepackage{color}
\begin{document}
\title{Are good local minima wide in sparse recovery?}
\author{\IEEEauthorblockN{Michael Moeller and Otmar Loffeld}
\IEEEauthorblockA{University of Siegen\\
H\"olderlinstra\ss e 3, 57076 Siegen}
\and
\IEEEauthorblockN{J\"urgen Gall}
\IEEEauthorblockA{University of Bonn\\
Endenicher Allee 19a, 53115 Bonn}
\and
\IEEEauthorblockN{Felix Krahmer}
\IEEEauthorblockA{Technical University of Munich\\
Boltzmannstr. 3, 85748 Garching b. M\"unchen}}
\maketitle

\newcommand{\MM}[1]{{\color{red}\textbf{#1}}}
\begin{abstract}
The idea of compressed sensing is to exploit representations in suitable (overcomplete) dictionaries that allow to recover signals far beyond the Nyquist rate provided that they admit a sparse representation in the respective dictionary. The latter gives rise to the sparse recovery problem of finding the best sparse linear approximation of given data in a given generating system. In this paper we analyze the iterative hard thresholding (IHT) algorithm as one of the most popular greedy methods for solving the sparse recovery problem, and demonstrate that systematically perturbing the IHT algorithm by adding noise to intermediate iterates yields improved results. Further improvements can be obtained by entirely rephrasing the problem as a parametric deep-learning-type of optimization problem. By introducing perturbations via dropout, we demonstrate to significantly outperform the classical IHT algorithm, obtaining $3$ to $6$ times lower average objective errors. 
\end{abstract}
\IEEEpeerreviewmaketitle

\section{Introduction}
\subsection{Sparse Recovery}
In many applications, the number of possible measurements is limited by physical or financial constraints on the measuring system. The field of compressed sensing has demonstrated that taking significantly fewer (linear) measurements than the number of unknowns still allows for exact recovery when assuming sparsity in a suitable (overcomplete) dictionary. By combining the measurement matrix and the dictionary into a single linear operator $A\in \mathbb{R}^{m \times n}$, the sparse recovery problem can be stated as the solution of the minimization problem 
\begin{align}
\label{eq:sparseRecovery}
\min_u \|Au-f\|^2 \qquad \text{s.t.} \quad |u|_0 \leq s,
\end{align}
for $f \in \mathbb{R}^m$ being measured data, $|u|_0$ denoting the number of nonzero entries in $u$, and $s$ being a known (or estimated) level of sparsity of the desired solution. 

Although \eqref{eq:sparseRecovery} is NP-hard in general (see e.g. \cite{npHard}), several greedy algorithms as well as convex relaxation approaches guarantee to find the global minimizer provided that $A$ meets certain regularity conditions such as the restricted isometry property (RIP), and often yield faithful approximate solution far beyond the theoretical guarantees, see \cite{Foucart13} for details.  

In general, a popular approach for solving constrained optimization problems of the form 
\begin{equation}
\label{eq:constrainedOpti}
\min_u E(u) \qquad \text{s.t.}  \qquad u\in M,
\end{equation}
for some set $M$ and an energy $E$ whose gradient $\nabla E$ is $L$-Lipschitz continuous, is to iteratively minimize a majorizer of $E$, giving rise to the so-called \textit{gradient projection} algorithm
\begin{equation*}
 u^{k+1} \in \arg\min_{u\in M} E(u^k) + \langle \nabla E(u^k), u-u^k \rangle + \frac{1}{2\tau}\|u -u^k\|_2^2, 
\end{equation*}
for $\tau \in ]0,\frac{1}{L}[$. Convergence follows from the majorization-minimization framework under weak assumptions, see e.g. \cite{Hunter04,Mairal13,Sun17}.  

Interestingly, although the set $M:= \{u\in \mathbb{R}^m~|~ |u|_0 \leq s\}$ is not convex, a (possibly not unique) projection can still be computed efficiently by simply keeping the $s$ largest entries in magnitude and setting the remaining entries to zero. Denoting this \textit{hard thresholding} operation by $H_s$, and applying the gradient projection algorithm to \eqref{eq:sparseRecovery} yields the \textit{iterative hard thresholding} (IHT) algorithm \cite{Blumensath2008,Blumensath2009},
\begin{align}
\label{eq:ihs}
u^{k+1} = H_s(u^k-\tau A^T(Au^k-f)). 
\end{align}
Despite its simplicity, the IHT algorithm is one of the most popular sparse recovery algorithms, and is among the greedy algorithms with the least restrictive assumptions for exact recovery (\cite[p. 29]{Foucart13}). 

\begin{figure}[t!]
\includegraphics[width=0.49\textwidth]{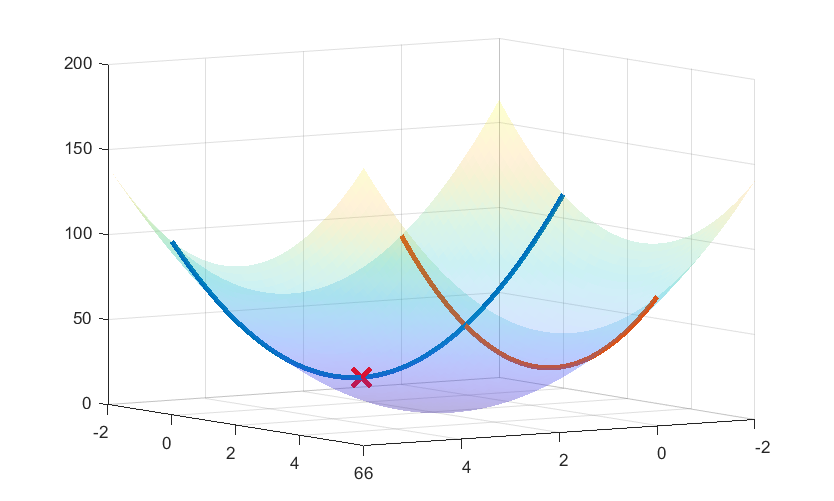}
\caption{Illustrating the energy landscape of \eqref{eq:sparseRecovery} for $m=2$ and $s=1$: The semi-transparent convex surface represents $\|Au-f\|^2$, which, however, is restricted to a feasible set of one-sparse solutions represented by the union of the orange and blue curves. While the red cross illustrates the global minimum, one can see that there is a local minimum on the orange curve, too. In this paper we investigate possible strategies to avoid getting stuck on such (inferior) local minima.}
\label{fig:energyLandscape}
\end{figure}

In general, we have to expect the energy landscape of \eqref{eq:sparseRecovery} to have many local minima on the feasible set, as we have illustrated for the simple case of $m=2$ and $s=1$ in Figure \ref{fig:energyLandscape}. Therefore, an interesting question is if there are systematic strategies that tend to avoid getting stuck in bad local minima?

Of course, one cannot expect a positive answer to this question for arbitrary nonconvex problems. However, let us assume/conjecture that good local minima tend to be wide for \eqref{eq:sparseRecovery} in many practically relevant settings, e.g. for random measurement matrices. As a consequence of the majorization-minimization framework, the IHT algorithm monotonically decreases the objective and provably converges to a local minimizer, see \cite{Blumensath2008}. If one perturbs the locally optimal solution several times and the IHT algorithm converges to the same locally optimal solution, one can conjecture to be in a wide local minimizer of the energy landscape, which - according to our assumption - has a comparably low energy. Vice versa, perturbing a locally optimal solution in a narrow minimum would likely lead to a different solution when continuing to optimize with IHT. 

With the above conjecture and intuition in mind, this manuscript investigates the behavior of two different strategies for obtaining perturbed IHT algorithms - one straight-forward algorithm that adds Gaussian noise to the IHT solution after a fixed number of iterations, and one algorithm that parameterizes the solution in terms of the IHT updates themself and adapts the idea of dropout \cite{dropout} from the world of deep learning.

After summarizing some related work in Section \ref{sec:related}, we present the perturbed IHT methods in Sections \ref{sec:nIHT} and \ref{sec:lIHT}. We demonstrate, in Section \ref{sec:numerics}, that the variants indeed exhibit systematically better performance as compared to the classical IHT  and draw (preliminary) conclusions in Section \ref{sec:conclusions}. 

\section{Related Work}
\label{sec:related}
\subsection{Sparse recovery}
As the central problem in the field of compressed sensing, the solution of \eqref{eq:sparseRecovery} has been studied in detail in the literature. While the exact solution to \eqref{eq:sparseRecovery} is known to be NP-hard to compute in general \cite{npHard}, many greedy strategies, e.g. orthogonal matching pursuit (OMP) \cite{omp}, CoSaMP \cite{cosamp}, the IHT algorithm \cite{Blumensath2008} discussed in the introduction, or hard thresholding pursuit (HTP) \cite{htp}, as well as $\ell^1$ minimizing convex relaxation approaches \cite{basisPursuit} are known to yield exact sparse recovery under certain conditions usually related to the restricted isometry property (RIP), see \cite{Foucart13} for an overview. Recent work on partial hard thresholding \cite{partialHT} provides a novel framework including new algorithms and nicely summarizes some recent exact recovery results. 

In this work we are less interested in exact recovery guarantees, but rather ask the question how close one can get to global minimizers of the nonconvex problem \eqref{eq:sparseRecovery} in practice. For this purpose we focus on perturbations of the IHT algorithm. Since we not only consider the straight-forward approach of adding noise to the iterates, but also a parametric approach based on deep learning ideas, we briefly summarize the main necessary concepts.

\subsection{Deep learning}
The core idea of deep learning is to use a parametric function $\mathcal{N}(x;\theta)$ that maps the input $x$ to the desired output. The \textit{network} $\mathcal{N}$ ideally has to be parameterized by the \textit{weights} $\theta$ in such a way that all desired outputs, but as few (undesirable) other elements as possible, lie in the range of the network.

The simplest and most generic architectures are \textit{fully connected networks}, for which $\mathcal{N}$ is given by a composition
\begin{equation}
\label{eq:mlp}
\mathcal{N}(x;\theta) = \phi(\sigma(\hdots \phi(\sigma(\phi(x;\theta_1));\theta_2)\hdots);\theta_L)
\end{equation}
of affine linear transfer functions
\begin{equation}
\label{eq:linearTransfer}
 \phi(x;\theta_i) = \theta_i^W x + \theta_i^b ,
\end{equation}
and nonlinearities $\sigma$, e.g. rectified linear units $\sigma(x) = \max(x,0)$. The overall parameters $\theta$ consist of the parameters $\theta_i$ of each \textit{layer}, which themselves typically divide into a weight matrix $\theta_i^W$ and a \textit{bias} $\theta_i^b$.

 Once an appropriate architecture for the network $\mathcal{N}$ has been chosen, the \textit{training} consists of an optimization problem 
\begin{equation}
\label{eq:learning}
 \hat{\theta} \in \arg\min_\theta \sum_{i \in \text{training set}} \mathcal{L}(\mathcal{N}(x_i;\theta),y_i) ,
 \end{equation}
in which one tries to determine parameters $\hat \theta$ that yield the lowest \text{loss} $\mathcal{L}$ on the \textit{training data}, typically consisting of pairs $(x_i,y_i)$ of inputs $x_i$ and desired outputs $y_i$ of the network. 
We refer the reader to \cite{deepLearningBook} for a more detailed introduction. 

To obtain a sufficiently expressive network many researchers have turned to deeply nested functions $\mathcal{N}$ with such a large number of weights $\theta$, that the training data can often be fitted almost perfectly. In order to regularize the training \eqref{eq:learning} of the network and prevent \textit{overfitting}, a very successful strategy is to introduce a \textit{dropout layer} \cite{dropout}, i.e. a random perturbation, into the training. More precisely, a dropout layer sets a certain fraction of the entries of an input vector to zero at random positions, and leaves the others untouched. This strategy has proven to be very efficient in preventing overfitting, and avoiding narrow local minima in the energy landscape of the training \eqref{eq:learning}, which makes it an interesting approach for our purposes. 

Finally, a common strategy for learning approaches inspired by variational or partial differential equation based approaches is to roll-out suitable algorithms and treat certain parts of the latter as learnable parameters of the resulting network -- see e.g. \cite{Zheng15,learningRDE, Richard17}
for details. 
\subsection{Dithering}
Another area where artificially introducing perturbations has proven useful is in analog-to-digital conversion, in this context this approach is known under the name of dithering. See for example \cite{XJ18} for a recent analysis of its benefits.
\section{Perturbing IHT}
\begin{figure}[t!]
\begin{center}
\includegraphics[trim={0cm 0.5cm 0cm 0.5cm}, clip,width=0.49\textwidth]{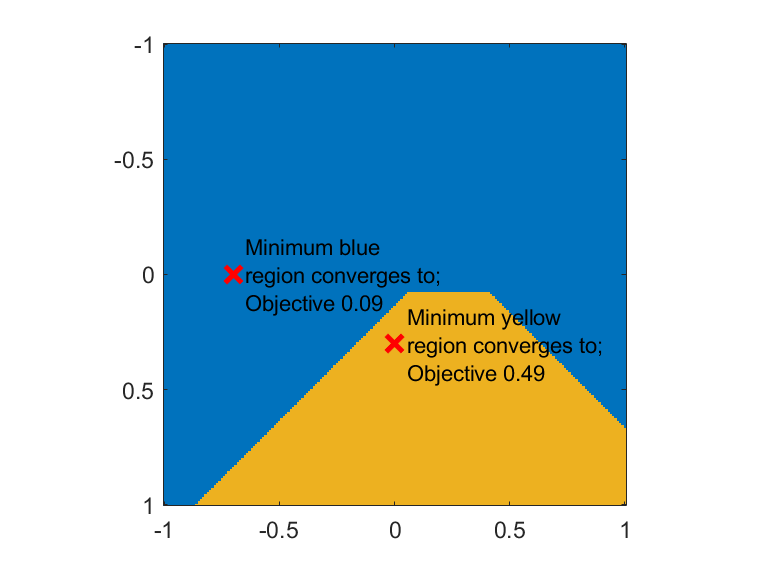}
\caption{Running IHT in a simple 2D setting often yields cases in which the region of starting points that converge to the global minimum is significantly closer to the suboptimal local minimum than the global minimum is to the region of points from which IHT converges to the suboptimal local minimum.}
\label{fig:2dExample}
\end{center}
\end{figure}

As an additional motivation for a perturbed IHT, we do the following test in 2D, i.e. for $n=m=2$ and $s=1$. We draw a random measurement matrix $A$ with columns of norm two, and random data $f$ with norm one. We run the IHT algorithm from $81\times 81$ different initial positions in $[-1,1]^2$ with step size $\tau = 0.05/\|A\|^2$. Figure \ref{fig:2dExample} illustrates the information such a run contains: One (typically) obtains two minima with corresponding regions in which one has to start to converge to the respective point. The conjecture the perturbed IHT algorithms are based on is that the suboptimal local minimum is closer to the region that converges to the global minimum than the global is to the region which converges to the suboptimal local one. The latter would make a perturbed algorithm more systematically more successful. While we picked a case where our assumption is met for illustration purposes in Figure \ref{fig:2dExample}, we ran the above random setting 1000 times, found 890 settings with two local minima in $[-1,1]$ and computed the average distances discussed above: Indeed the average distance of the global minimum to the local region was 0.254 while the distance of the suboptimal local minimum to the global region was 0.144. Of course, the above test is very simplistic and we had to pick a relatively small $\tau$ in order to arrive local minima more often. Nevertheless, we hope for the intuition to carry over to the significantly more difficult recovery problems in higher dimensions. The following section discusses two variants of perturbed IHT algorithms. 

\subsection{Adding noise}
\label{sec:nIHT}
The first (straight-forward) way to perturb the IHT algorithm is to add noise after a fixed number of iterations. For this purpose, we run the IHT algorithm for 600 iterations before adding zero-mean Gaussian noise with standard deviation $\sigma = 0.025$ to its outcome, which is subsequently used as the input to another 600 IHT iterations. We stop the process after the fifth run of the IHT algorithm. While this procedure can surely be optimized in terms of the hyperparameters, the number of times the system is perturbed, and the tracking of the best solution among all reinitializations, the goal of our work is to see if this simple perturbation strategy allows to demonstrate a systematic improvement over the plain IHT algorithm (which we run for $5\cdot 600=3000$ iterations for the sake of fairness in our numerical experiments). We refer to this variant as the \textit{noisy IHT algorithm}. 

\subsection{Optimizing the algorithm with dropout}
\label{sec:lIHT}
An entirely different approach is roll out the IHT algorithm and treat the updates themselves as learnable parameters: Note that $L$ iterations of the IHT algorithm can be written as
\begin{equation}
\label{eq:rolledOutIht}
u^L = H_s(\phi( \hdots H_s(\phi(H_s(\phi(u^0;\theta_1));\theta_2))\hdots ; \theta_L)),
\end{equation}
for $\phi$ denoting the linear transfer functions defined in equation \eqref{eq:linearTransfer}, and the parameters $\theta_i$ taking the specific form 
\begin{equation}
\label{eq:paramInit}
\theta_i^W = Id - \tau A^TA, \quad \theta_i^b = \tau A^Tf, \quad \forall i\in \{1,\hdots,L\}.
\end{equation}
In full analogy to \eqref{eq:mlp} one could now treat $u^L =: \mathcal{N}(u^0;\theta)$ as a neural network 
and try to optimize
\begin{equation}
\label{eq:optimization}
\min_\theta \|A\mathcal{N}(u^0;\theta) -f \|_2^2. 
\end{equation}
Note, however, that using \eqref{eq:paramInit} as an initialization for the weights along with a deep network leads to the initial parametrization yielding a local minimizer such that gradient based weight optimization can never progress. Moreover, each layer of the parametrization \eqref{eq:rolledOutIht} has $n(n+1)$-many free parameters, making it expensive to optimize deep architectures. 

We therefore fix the first 3000 iterations to the noisy IHT algorithm and merely use two further IHT iterations as our parametrization. To avoid the problem of starting at a local minimum, we pick up the perturbation idea and introduce an intermediate dropout layer. More precisely, we consider the architecture
\begin{equation}
\label{eq:proposed}
\mathcal{N}(u^0; \theta) = H_s(\phi(H_s(\text{drop}(\phi(u^0;\theta^1);5\%));\theta^2)),
\end{equation}
use the result of the noisy IHT algorithm as $u^0$, initialize according to \eqref{eq:paramInit}, and optimize \eqref{eq:optimization}. The layer $\text{drop}(x; 5\%)$ denotes randomly setting $5\%$ of the entries of $x$ to zero and is only used during the optimization, not for the final prediction. We refer the reader to \cite{dropout} for details on such a \textit{dropout}. 

For the optimization itself we utilize a subgradient-descent-type algorithm with momentum as commonly used in learning applications. Similar to the way rectified linear or max-pooling units are optimized in the deep-learning literature, we ignore the set of non-differentiable points in the hard-thresholding operator $H_s$ and use
$$ (\nabla H_s(x))_i = (|(H_s(x))_i|>0)\cdot x_i $$
as a (sub-)derivative. Practically, we use Matlab's deep learning framework, set the momentum to $0.9$, the step size (learning rate) to $10^{-4}$, and run the algorithm for 2000 iterations. Despite the lack of convergence guarantee for such an algorithm, it successfully reduced the objective value in all test cases. We refer to this approach as the \textit{parametric IHT method}. 

Note that the parametric IHT method is not a learning based approach to sparse recovery, but merely uses a particular parametrization of the solution space which exploits some concepts borrowed from the deep learning literature. Similar ideas of regularization by reparametrization have recently been made for image reconstruction in \cite{deepImagePrior}, where -- similar to the idea of \eqref{eq:optimization} -- a deep network based architecture is used to parameterize natural images. 

It is worth pointing out that the bias of the last linear transfer function is of the same size as the desired solution already, such that \eqref{eq:proposed} represents a significant overparametrization, which on the other hand also appears to be the reason why the method outperforms classical optimization algorithm such as IHT significantly, as we will see in the next section. Similar to other areas of nonconvex optimization, e.g. \cite{Zach14}, increasing the dimensionality of the underlying problem seems to help in avoiding local minima even before entering a setting of convex lifting approaches.

\begin{figure*}[tbh]
\includegraphics[width=0.32\textwidth]{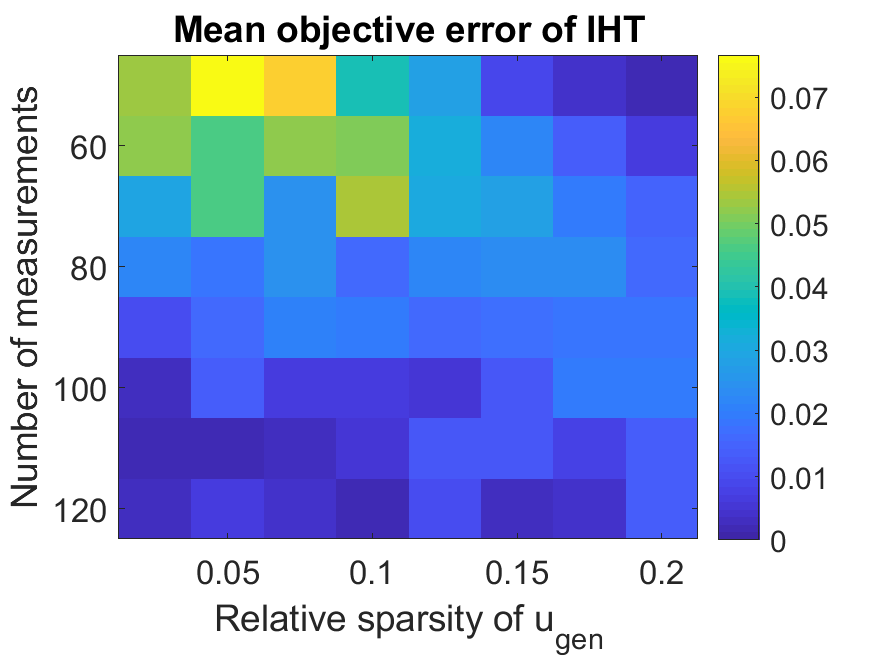}
\includegraphics[width=0.32\textwidth]{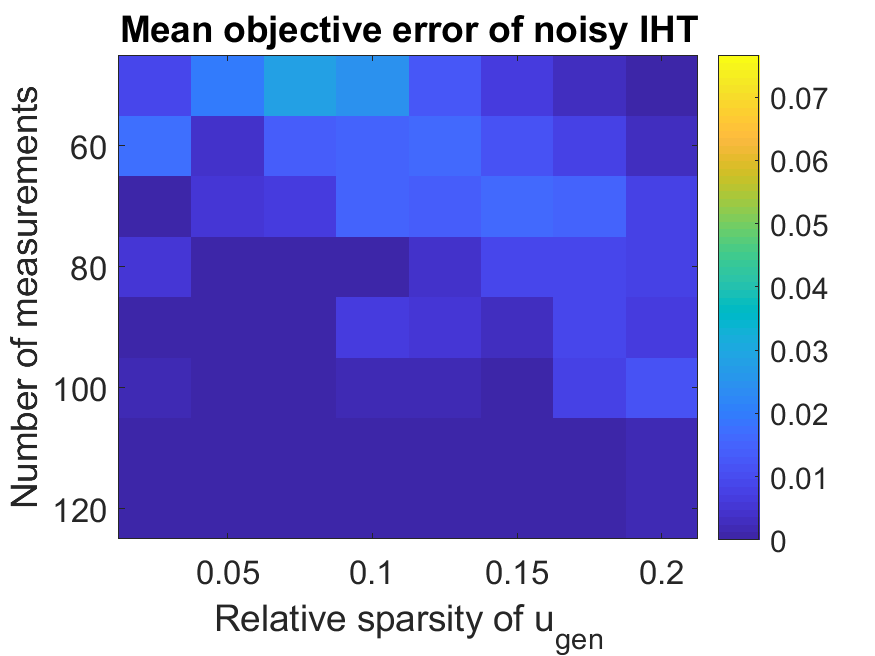}
\includegraphics[width=0.32\textwidth]{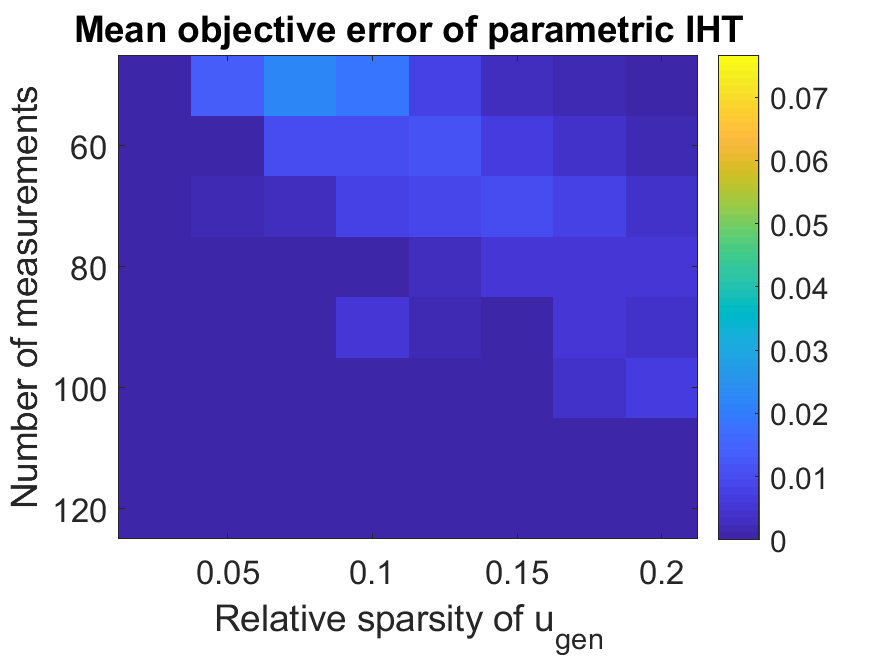} \\
\centering Mean of $\|Au-f\|^2$ over all 20 runs for each method. \\ 
$ $ \\
\includegraphics[width=0.32\textwidth]{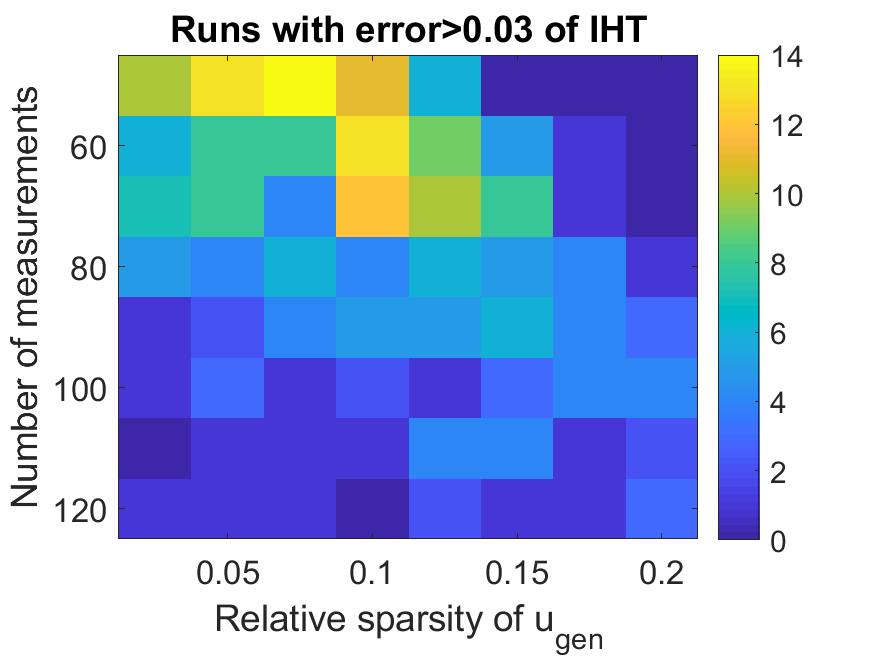}
\includegraphics[width=0.32\textwidth]{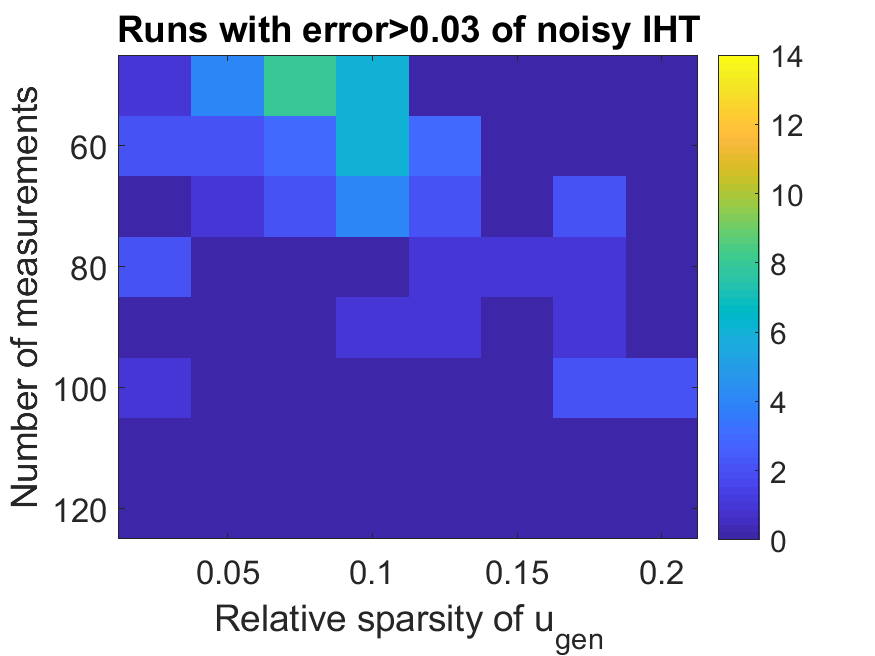}
\includegraphics[width=0.32\textwidth]{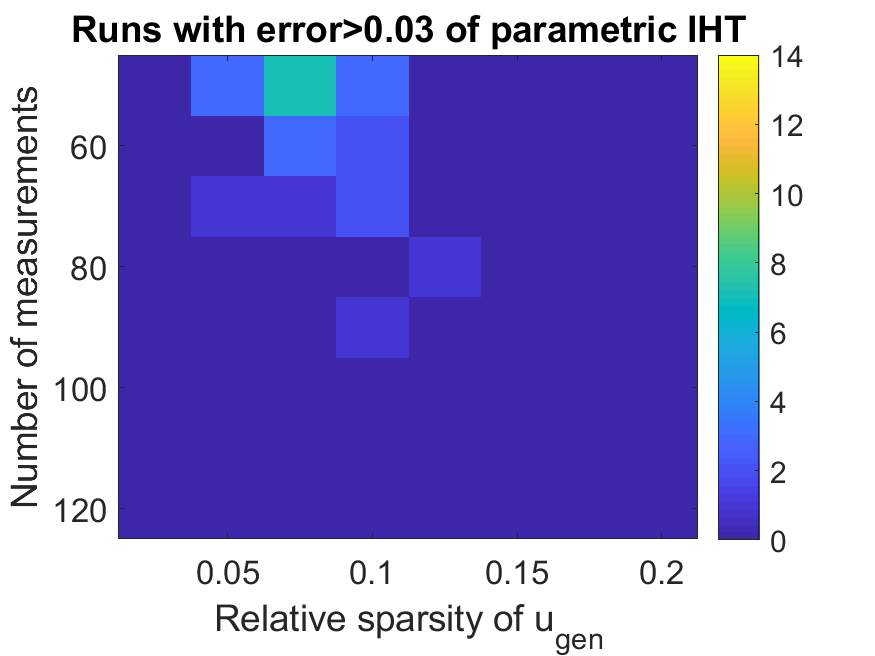} \\
\centering Count how often $\|Au-f\|^2>0.03$ for each method within the 20 runs. \\ 
$ $ \\
\includegraphics[width=0.32\textwidth]{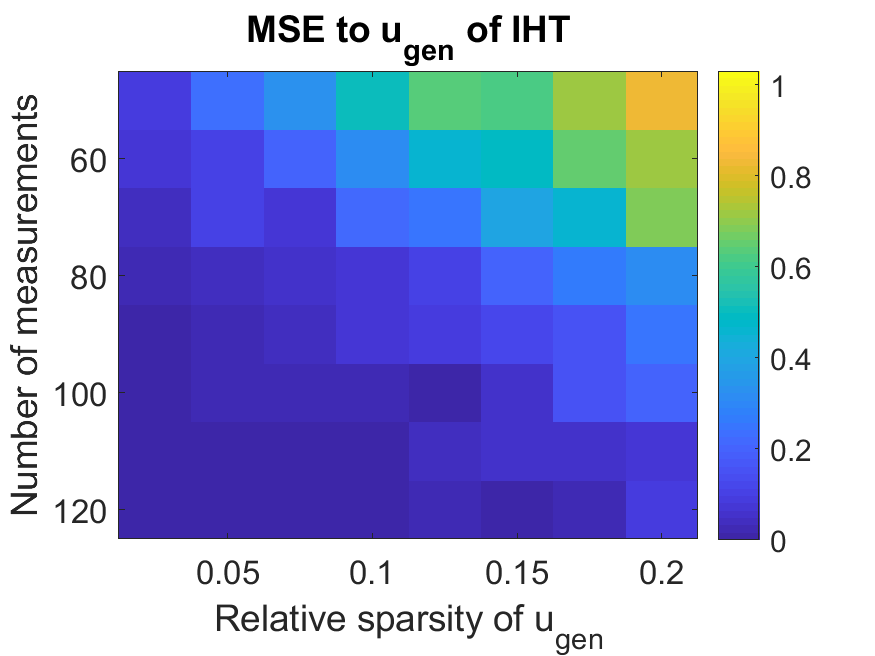}
\includegraphics[width=0.32\textwidth]{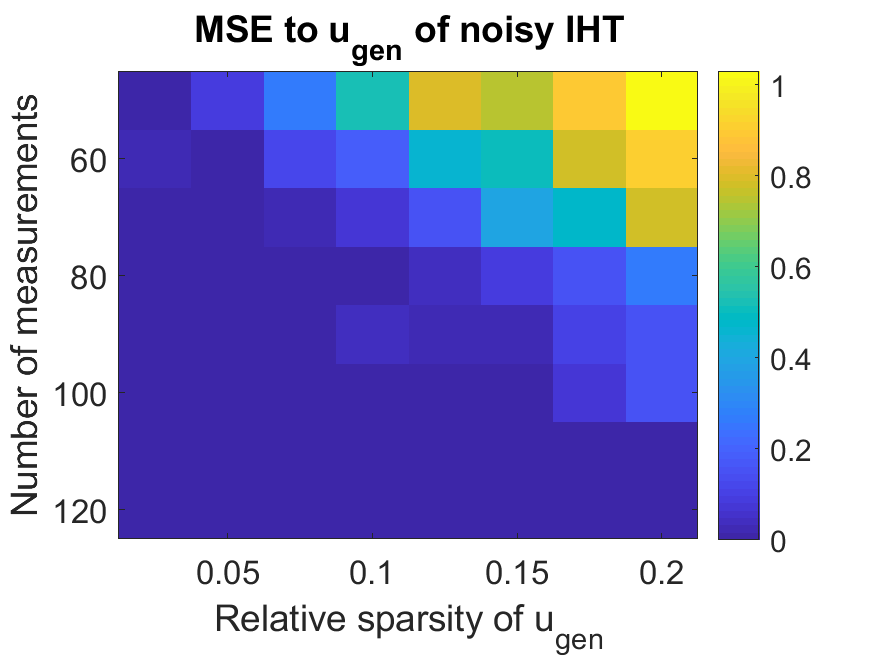}
\includegraphics[width=0.32\textwidth]{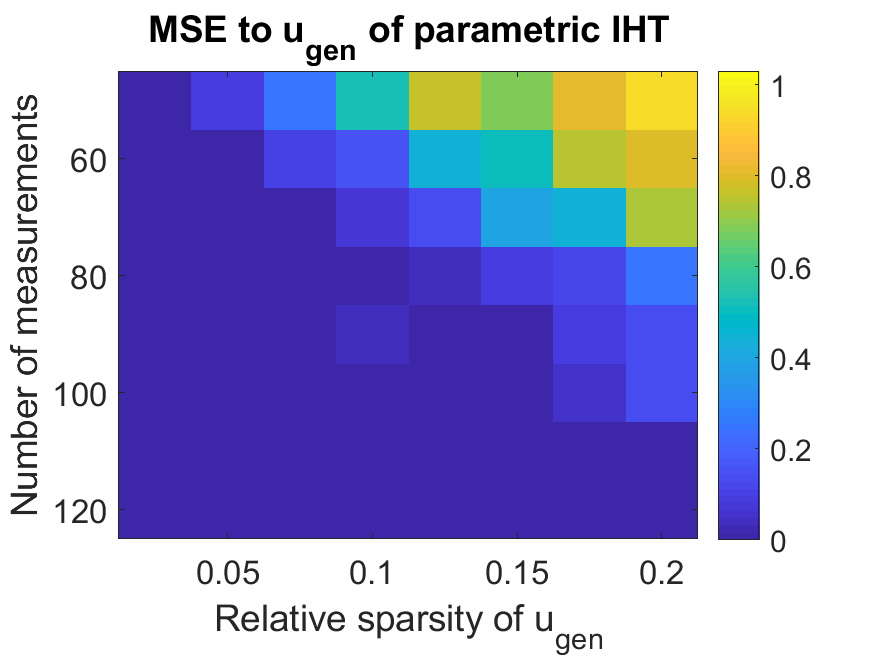} \\
\centering  Mean of $\frac{\|u-u_{gen}\|^2}{\|u_{gen}\|^2}$ for each method over the 20 runs.\\
\caption{Comparison of the average objective error $\|Au-f\|^2$ (first row), the number of times the objective error exceeded $0.03$ among 20 runs (middle row), and the relative difference $\|u-u_{gen}\|^2/\|u_{gen}\|^2$ to the element that generated the data $f=Au_{gen}$.}
\label{fig:comparison}
\end{figure*}

\section{Numerical results}
\label{sec:numerics}
We investigate the behavior of the IHT, the noisy IHT, and the parametric IHT algorithms by generating matrices $A\in \mathbb{R}^{m\times n}$ with entries sampled from a Gaussian distribution, using a fixed $n=200$, and varying $m\in \{50,60,70,\hdots,120 \}$. We generate data $f=Au_{gen}$ for $u_{gen}$ having $s=\text{round}(\mu \cdot n)$ many nonzero entries also drawn from a Gaussian distribution. We vary the relative sparsity $\mu \in \{0.025,0.05,0.075,\hdots, 0.2\}$, and finally normalize to $\|f\|_2=1$. We then run each of the algorithms with the number $s=|u_{gen}|_0$ of nonzero entries we used to generate the data with, and measure their performance in terms of the objective error $\|Au-f\|^2$. To avoid the issue of possibly having taken too few iterations, we solve one final least-squares problem on the support of the solution returned by each of the algorithms. 

The first row of figure \ref{fig:comparison} shows the average objective error over 20 realizations of the IHT, the noisy IHT, and the parametric IHT approaches. The second row shows the count how often (among the 20 overall runs) the objective error exceeded a value of $0.03$, and, for the sake of interpretability, the third row shows the normalized difference to the generating element $\|u-u_{gen}\|^2/\|u_{gen}\|^2$ for each of the three methods. 

We conducted similar experiments with measurement matrices $A$ arising from a Bernoulli-sampling, and subsampled discrete cosine transform matrices. Since the results were qualitatively similar, we are merely showing the Gaussian case for the sake of brevity. 

As an additional summary, Table \ref{tab:comparison} shows the average objective error of all three methods over all experiments of Figure \ref{fig:comparison}. 
As we can see, both perturbations of the IHT algorithm led to significant improvements over the classical IHT with the overall average objective error being improved by a factor of $3.4$ by the noisy IHT and by a factor of $6$ by the parametric IHT. While the number $2n(n+1)$ of free parameters of the parametric IHT approach is notably larger than the $n$ parameters of the original unknown (leading to a runtime of about 7-8 seconds on a CPU in the above test), the resulting optimization with dropout is significantly less prone to getting stuck in local minima and yields surprisingly large improvements. 

\newpage
As we can see in Figure \ref{fig:comparison}, a small support of $u_{gen}$ along with a small number of measurements caused relatively large objective errors of the classical IHT, which could be improved significantly by its perturbed variants. Since the histogram graphs (second row in Figure \ref{fig:comparison}) look quite similar to the mean objective errors of the first row, we can conclude that the errors of IHT are not just based on a single unlucky random realization, but yield systematic problems with up to $14$ failures to reduces the objective below $0.03$. 

Note that the goal of this work is to investigate the minimization of $\|Au-f\|$ under sparsity constraints rather than the exact recovery of the generating element $u_{gen}$. Nevertheless, we included the normalized distance to $u_{gen}$ for the sake of completeness in the last row of Figure \ref{fig:comparison}. We would, however, like to point out that settings with few measurements and large supports, e.g. $m=50$ and $s=40$ (upper right part of the plots in the last row of Figure \ref{fig:comparison}), naturally imply that we can neither expect $u_{gen}$ to be the unique element to meet $f=Au$, nor the sparsest. This means that although the noisy and parametric IHT gave very high values of $\|u-u_{gen}\|^2/\|u_{gen}\|^2$ in such cases, they do provide desirable solutions with almost no objective errors. Moreover, differing from $u_{gen}$ in only one or two components can already lead to comparably high objective errors in the case of few measurements as we can see in the IHT graphs in the upper left. Nevertheless, in settings of sparse $u_{gen}$ and moderately many measurements, the last row of Figure \ref{fig:comparison} also indicates that the perturbed IHT algorithm have a higher success rate in recovering $u_{gen}$.

\begin{table}[tb]
\caption{Average objective error $\|Au-f\|^2$ of the classical IHT, the noisy IHT, and the parametric IHT algorithms, over all relative sparsities and all number of measurements from Figure \ref{fig:comparison}.}
\label{tab:comparison}
\def\arraystretch{1.5}%
\begin{center}
\normalsize
\begin{tabular}{l|ccc}
Method & IHT & Noisy IHT & Param. IHT \\\hline
Avg. $100  \cdot \|Au-f\|^2$&$2.0$ & $0.59$& $0.33$ 
\end{tabular}
\end{center}
\end{table}

Future research will try to give some theoretical explanations for the behavior we observed in the the numerical studies above. In particular, we will consider the case of noisy measurements and investigate strategies to quantify by what amount the IHT algorithm should ideally be perturbed. Furthermore, an interesting question is the optimal design of the network-like parametrization.

\section{Conclusion}
\label{sec:conclusions}
In this paper we have considered the nonconvex optimization problem of minimizing a quadratic objective subject to a constraint on the maximum number of nonzero entries of the unknown. Based on the assertion that good local minima are wide, we compared the IHT algorithm with a simple modification that perturbs the iterates with noise, as well as with a reparametrization of the solution in a network-like architecture which was perturbed by incorporating dropout. It was shown that the noisy IHT yields significant improvements over the plain IHT algorithm, while the parametric IHT approach involving dropout can further improve the noisy IHT results by a large margin. In summary, our findings support the hypothesis that good local minima are wide for randomly sampled measurement matrices and exact data, and demonstrate that a dropout-based parametrization of the solution space is an excellent approach for exploiting this property. 

\bibliographystyle{IEEEtran}
\bibliography{refs}
\end{document}